\documentclass[letterpaper, 10 pt, conference]{ieeeconf}  

\IEEEoverridecommandlockouts                              

\usepackage{xcolor}
\usepackage{tablefootnote}

\title{\LARGE \bf
Residual Feedback Learning for Contact-Rich Manipulation Tasks with Uncertainty
}

\author{Alireza Ranjbar$^{1,2,3}$
\and Ngo Anh Vien$^{3}$
\and Hanna Ziesche$^{3}$
\and Joschka Boedecker$^{1}$
\and Gerhard Neumann$^{2}$
\thanks{Affiliations with}
\thanks{$^{1}$ Albert-Ludwigs-Universität Freiburg} \thanks{ $^{2}$ Karlsruhe Institute of Technology} \thanks{$^3$ Bosch Center for Artificial Intelligence (BCAI)}
}

\renewcommand{\baselinestretch}{.96}
\renewcommand{\textfloatsep}{1ex}

\usepackage[acronym]{glossaries}
\makeglossaries

    \newacronym{ml}{ML}{machine learning}
    \newacronym{rl}{RL}{reinforcement learning}
    \newacronym{drl}{DRL}{deep reinforcement learning}
    \newacronym{mbrl}{RL}{model based reinforcement Learning}
    \newacronym{lfd}{LfD}{learning from demonstration}
    \newacronym{mpc}{MPC}{Model Predictive Control}
    \newacronym{rpl}{RPL}{residual policy learning}
    \newacronym{rfl}{RFL}{residual feedback learning}
    \newacronym{pearl}{PEARL}{Probabilistic embeddings for actor-critic RL}
    \newacronym{ffnn}{FFNN}{Feed forward neural networks}
    \newacronym{bo}{BO}{Bayesian Optimization}
    \newacronym{mdp}{MDP}{Markov decision process}
    \newacronym{pomdp}{POMDP}{Partially observable Markov decision process}
    \newacronym{spd}{SPD}{symmetric positive definite}
    \newacronym{via}{VIA}{variable impedance actuators}
    \newacronym{cmaes}{CMA-ES}{Covariance Matrix Adaptation Evolutionary Strategy}
    \newacronym{ppo}{PPO}{proximal policy optimization}
    \newacronym{trpo}{TRPO}{Trust region policy optimization}
    \newacronym{lqr}{LQR}{linear quadratic regulator}
    \newacronym{lqg}{LQG}{Linear quadratic Gaussian}
    \newacronym{gp}{GP}{Gaussian processes}
    \newacronym{mba}{MBA}{Model based acceleration}
    \newacronym{gps}{GPS}{Guided policy search}
    \newacronym{mdgps}{MDGPS}{mirror descent guided policy search}
    \newacronym{mve}{MVE}{Model-Based Value Expansion}
    \newacronym{rnn}{RNN}{Recurrent neural networks}
    \newacronym{lstm}{LSTM}{long short term memory}
    \newacronym{gru}{GRU}{Gated recurrent unit}
    \newacronym{kl}{KL}{Kullback-Leibler}
    \newacronym{gae}{GAE}{generalised advantage estimation}
    \newacronym{mp}{MP}{Manipulation primitive}
    \newacronym{ddpg}{DDPG}{Deep Deterministic Policy Gradient}
    \newacronym{her}{HER}{Hindsight experience replay}
    \newacronym{lhs}{LHS}{Latin-hyper-cube sampling}
    \newacronym{dof}{DOF}{degrees of freedom}
    \newacronym{tcp}{TCP}{tool central point}

\usepackage{graphicx}
\usepackage{subfig}
\usepackage{multirow}
\usepackage{array}
\usepackage{float}
\usepackage{amsfonts}
\usepackage[hidelinks]{hyperref}
\begin{document}

\maketitle
\thispagestyle{empty}
\pagestyle{empty}

\begin{abstract}


 While classic control theory offers state of the art solutions in many problem scenarios, it is often desired to improve beyond the structure of such solutions and surpass their limitations. To this end, \emph{\gls{rpl}} offers a formulation to improve existing controllers with reinforcement learning (RL) by learning an additive "residual" to the output of a given controller. However, the applicability of such an approach highly depends on the structure of the controller. Often, internal feedback signals of the controller limit an RL algorithm to adequately change the policy and, hence, learn the task.  We propose a new formulation that addresses these limitations by also modifying the feedback signals to the controller with an RL policy and show superior performance of our approach on a contact-rich peg-insertion task under position and orientation uncertainty. In addition, we use a recent Cartesian impedance control architecture as the control framework which can be available to us as a black-box while assuming no knowledge about its input/output structure, and show the difficulties of standard RPL. Furthermore, we introduce an adaptive curriculum for the given task to gradually increase the task difficulty in terms of position and orientation uncertainty. A  video  showing  the  results  can  be found at 
 \href{https://youtu.be/SAZm_Krze7U}{\texttt{https://youtu.be/SAZm{\char`_}Krze7U}.}


\end{abstract}

\section{INTRODUCTION}

Humans' skills for manipulating their environment has historically been foreseen as being overtaken by machines for which recent decades of research in artificial intelligence have promised more dexterity, adaptability, and cost efficiency with the help of accumulated experience or data. On the frontier, \gls{drl} has proven its capability at learning similar skills through data and prior-knowledge, offering novel solutions while often surpassing humans' performance similar to other advances in \gls{ml}. However, as the research in this field continues, major challenges such as sample complexity and generalization capacity of the algorithms are yet addressed differently given each problem scenario and in many cases no solution is known to be optimal.

\begin{figure}
\centering
\vspace{-0.1pt}
\subfloat[Hybrid Residual Reinforcement Learning]{\includegraphics[width=0.486\textwidth]{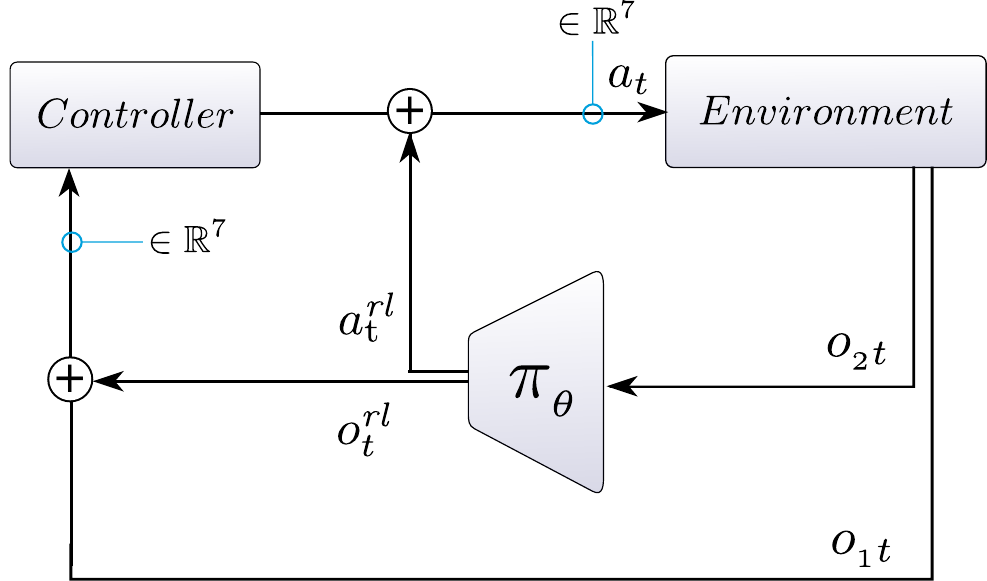}}
\qquad   

\subfloat[Expected behaviour of each Residual policy formulation]{%
        {\includegraphics[width=0.475\textwidth]{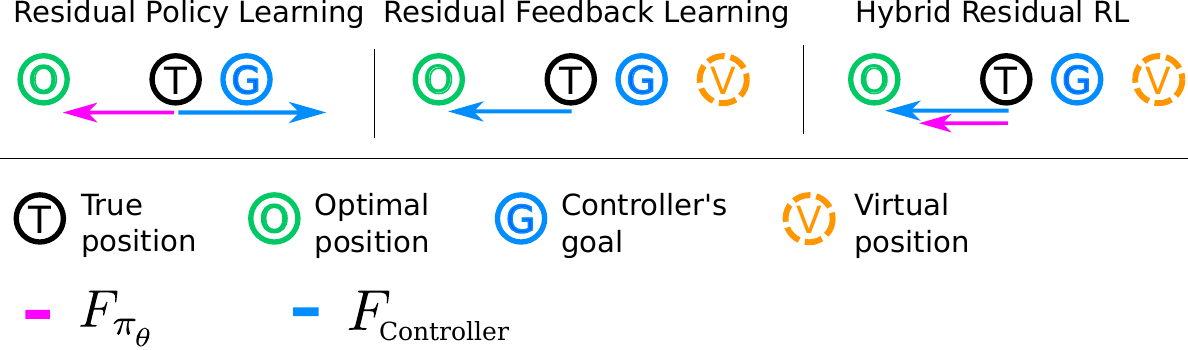}}
        }%
\caption{(a) RPL combined with RFL. Notations $o_{_1{t}}$, $o_{_2{t}}$, $a_{t}^{rl}$, and $a_t$, correspond to the observations of the prior controller, and the RL policy, inferred actions from the RL policy, and final actions applied to the environment. 
(b) While RPL regards the intervention of the RL policy as external perturbation and error compared to its goal, it resists the intervention. In contrast, RFL changes the goal itself through feedback, that is for example, the controller sees a virtual position instead of the true position.}
\label{fig:res_policies_expectation}
\end{figure}

In many problem settings improving over available solutions appears more applicable than learning skills from scratch. Especially when samples are expensive, more guarantees are necessary, or the need for a solution is perceived more crucial than discovery while engineering costs are undesirable. Residual policy learning (RPL) \cite{res_rl}, as one of possible solutions in this regard, suggests a formalism where the \gls{rl} agent learns to compensate for the imperfections of the up-stream controller by superposing its actions with it. This formulation provides a trade-off between capturing more information from the environment or allowing more exploitation of prior knowledge. Yet, the up-stream controller in many cases sees the intervention of the \gls{rl} policy as external perturbation and error, and therefore tries to resist it. Fig. \ref{fig:res_policies_expectation} (b) illustrates an example on the left where an ideal \gls{rl} policy applies force toward the optimal direction while the controller has a different goal. In this case, one can instead modify the feedback to the controller and obtain different results. In this formulation which we denote as "residual feedback learning" (RFL) the controller observes, for example, a \emph{virtual} position, instead of the true feedback, and is therefore  promoted in a different direction. In addition, in places where each formulation has its own advantage, combining both, i.e. an RL policy that outputs the residual controls as well as residual feedback commands, allows leveraging the distinct advantages of both methods simultaneously in one framework.

In general, for tasks that require wide spatial movements, e.g, moving toward the hole in a peg insertion task, RPL appears limited as the up-stream controller observes the external intervention of an ideal RL policy as error from its goal and tries to recover from it. In contrast, RFL causes the controller to observe a virtual feedback (e.g., position) instead of the true feedback and hence does not result in a competition between the RL policy and the controller. On the other hand, for tasks that require sudden actions or high frequency vibrations such as releasing a stuck peg due to the orientation uncertainty of the hole, RFL does not appear suitable as the up-stream controller is often designed to only have smooth outputs while filtering feedbacks in different ways. In contrast, in this scenario RPL can apply such sudden actions. For this reason, we want to leverage the distinct advantage of each formulation at the same time. We refer to this approach as Hybrid Residual Reinforcement Learning (HRRL) and illustrate in our experiments the beneficial performance of such approach. 

\begin{figure*}
  \begin{tabular}{c}
    \subfloat[joint effort action]{\includegraphics[width=0.245\textwidth]{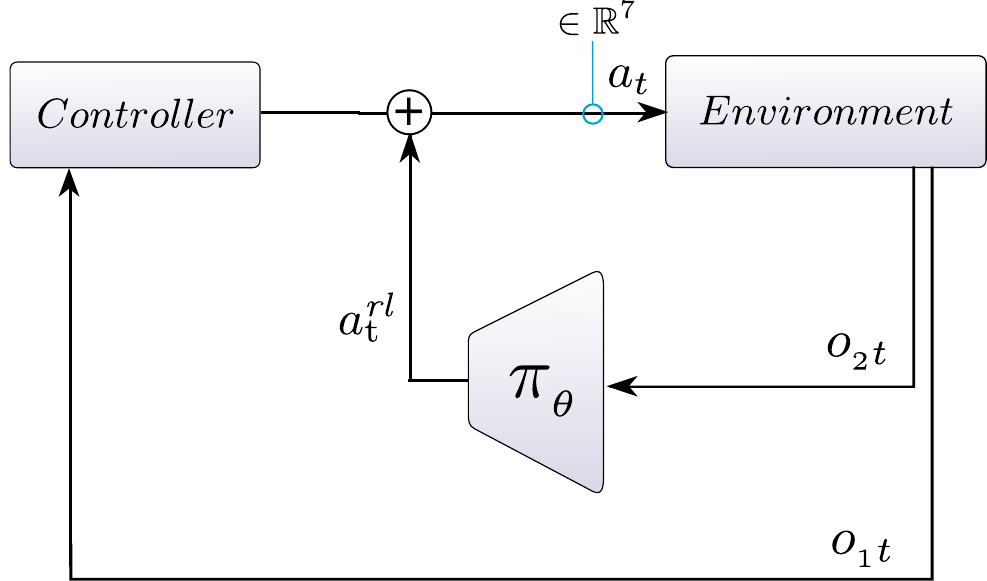}} \\
    \subfloat[end-effector wrench]{\includegraphics[width=0.245\textwidth]{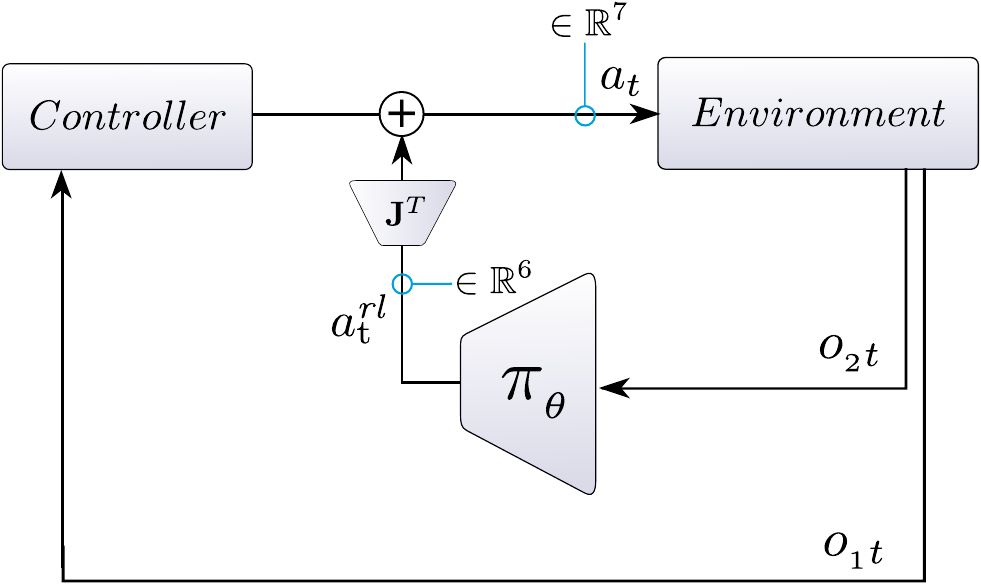}}
  \end{tabular} \hfill
  \begin{tabular}[m]{c}
    \subfloat[State machine behavior sequence]{\includegraphics[width=0.40\textwidth]{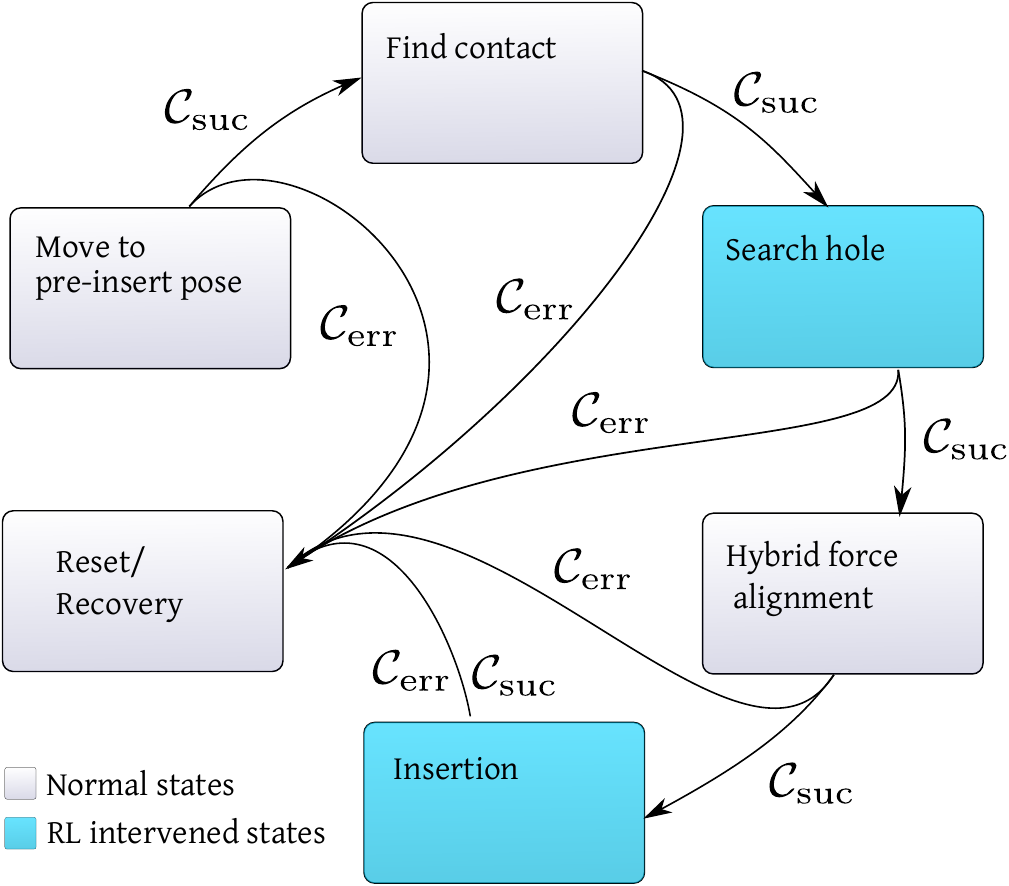}}
  \end{tabular} \hfill
  \begin{tabular}{c}
    \subfloat[joint position feedback]{\includegraphics[width=0.245\textwidth]{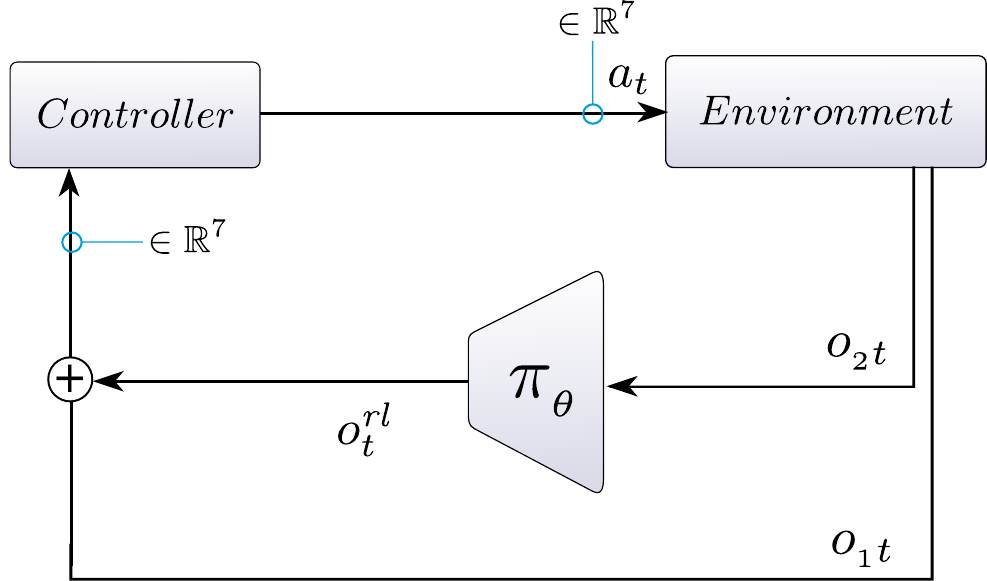}} \\
    \subfloat[end-effector pose feedback]{\includegraphics[width=0.245\textwidth]{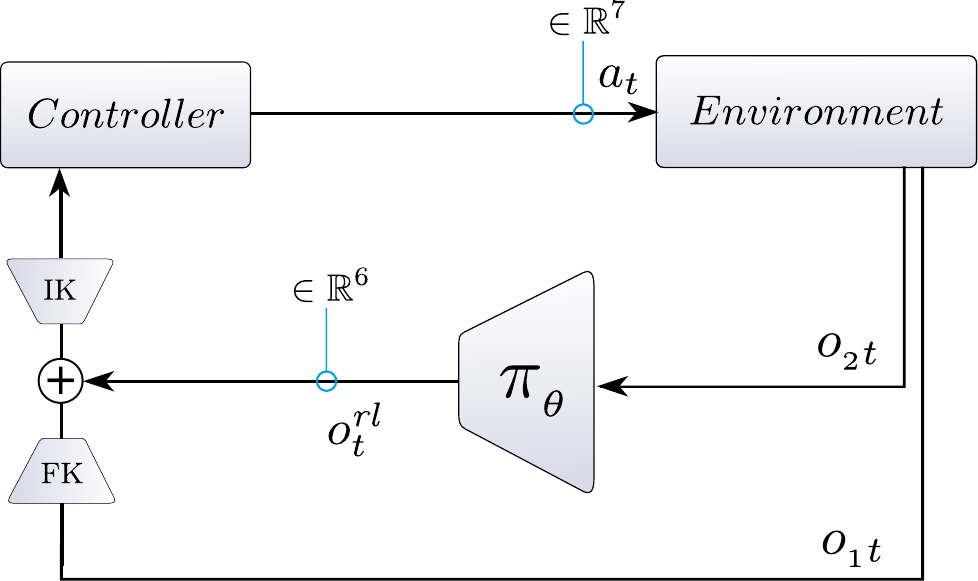}} 
  \end{tabular}
\caption[Residual policy variants]{(c): State machine behavior sequence of our insertion formalism. (a): The vanilla residual policy formulation from \cite{res_rl, silver2018residual}. (b): An alternative to allow inferring residual wrench values instead of residual joint torques. (d): The vanilla residual feedback formulation in contrast to (b). (e): Similar to (b) it is an alternative allowing to infer residual feedback in task space instead of joint space, e.g., the end effector pose. We elaborate more details regarding each in section \ref{sec:method}.}%
 \label{fig:res_policies_vars}%
\end{figure*}

Furthermore, we build our residual reinforcement learning algorithm on the recently developed manipulation framework of \cite{imped_meta} which provides an adaptive and compliant controller based on impedance control for several manipulation tasks.  Using this baseline, we show the applicability of our formulation for industrial assembly environments represented by the common challenging peg-in-the-hole task. In addition, this approach allows having the option to choose where the RL policy should intervene for improvement as shown in Fig. \ref{fig:res_policies_vars}. Finally, we evaluate and compare different variations of our residual policies for each sub-task of peg-in-the-hole, as well as the complete task in simulation. We considerably increase the task complexity by adding significant uncertainty in position and orientation of the hole, rendering it impossible for the standard controller to overcome. In order to cope with this challenging scenario, we apply adaptive curriculum learning to vary the task difficulty in terms of these uncertainties, leading to significant performance improvements.

Our primary contributions are as follows:
\begin{itemize}
    \item We propose an alternative and an extension to the \gls{rpl} formulation \cite{res_rl} to address its limitations for a wide range of tasks and controllers.
    \item We extend the manipulation framework of \cite{imped_meta} using our approach and illustrate how the addition of residual feedback removes some limitations of standard residual policy learning. 
    \item An empirical evaluation of the original method and ours along with their variants in simulation that we train within a recently proposed adaptive curriculum formalism \cite{lukas}.

\end{itemize}

\section{RELATED WORK}
\paragraph{Residual Reinforcement Learning}
 Two concurrent works \cite{res_rl} and \cite{silver2018residual} demonstrated the \gls{rpl} formulation and highlighted advantages such as sample efficiency, better sim-to-real adaptation, as well as the ability in handling sensor noise and controller miscalibration. A follow-up work \cite{schoettler2019deep} developed this idea further using visual inputs and sparse rewards for industrial insertion tasks. Other work investigated improving the performance of RPL by exploiting the uncertainty of the policy architecture to decide when only the bare controller should be used \cite{rana2019residual} or taking advantage of using more than one controller \cite{barekatain2019multipolar}.

\paragraph{Contact Rich Manipulation and Assembly}

Early works regarding peg-in-the-hole insertion had a rather theoretical view for analyzing contact models between the peg and the hole \cite{earlyPIH2,ding2019transferable}. A number of works focused on task specific engineering efforts or obtaining an accurate state estimation of contact through analytical or statistical methods \cite{earlyPIH4}. On the other hand, some of the learning-based approaches include \gls{lfd} \cite{LfD_PIH_1}, model-free \gls{rl} with proprioceptive and/or visual feedback \cite{levine2016endtoendviso,force2control,lee2019making}, model-based RL \cite{levine2015ContactRichManipul,abbeel_manipul18}, and meta-RL \cite{schoettler2020metareinforcement}. A concurrent work \cite{hoppe_20} also leveraged residual-\gls{rl} for an insertion problem. However, the authors mainly focused on analyzing the performance of a newly proposed graph-based structure for the experience replay buffer used commonly in off-policy \gls{rl} methods. There has been recent effort proposing to take the advantages of the complementary nature of both haptic and visual inputs for industrial manipulation tasks \cite{hoof_RL_AE_Haptiv_vision,inoue2017deep,lee2019making,haugaard2020fast}. The formulations we discuss and propose in this work can certainly leverage the above ideas as well, as they remain agnostic to the choices of state representation, policy architecture, and the training algorithm.

\paragraph{Impedance Control} Variable impedance actuators (VIA) offer various natural characteristics of human motion such as safety, robustness, and energy efficiency while still possessing fast response time to impacts as well as energy efficiency \cite{DLR_LWR_07, imped_controller_11}. A recent work that applied VIA \cite{imped_controller_11} proposed the first controller that can simultaneously adapt force and impedance within unknown dynamics to handle unstable conditions without requiring sensation of interaction forces. 
This work was then extended by Johannsmeier et. al. \cite{imped_meta} to Cartesian space and full feed-forward tracking to also offer a structure for Cartesian impedance control that is applicable in a variety of tasks. Accordingly, the authors exploit the knowledge regarding constraints that come with every hardware such as stiffness adaptation speed. Furthermore they define a graph based manipulation skill formalism that can reduce the complexity of the solution space for robots' force-sensitive manipulation skills. We leverage these ideas in this work, while in contrast, we resort to a finite state machine controller.

\section{Problem Statement}
\subsection{Partially Observable Markov Decision Process}
We assume a controller is already available over which we aspire to improve using \gls{rl}. Similar to most \gls{rl} works in manipulation skills that involve uncertainty, we also formulate our problem as a discrete time and episodic \gls{pomdp} described by the tuple $\mathcal{M}={(\mathcal{S}, \mathcal{A}, \mathcal{O}, \mathcal{P}, \mathcal{E}, \mathcal{R}, \gamma)}$. These entries respectively correspond to the state-space, action-space, observation-space, transition probability $\mathcal{P}\left(s_{t+1} \mid s_{t}, a_{t}\right)$, emission probability $\mathcal{E}\left(o \mid s\right)$, reward function $r(s, a)$, and discount factor $\gamma$; where $ s\in \mathcal{S},a\in \mathcal {A},o\in \mathcal{O}$. 
We also define  $R\left(\tau\right)=\sum_{i=t}^{T} \gamma^{i-t} r(s_{i}, a_{i})$ as our discounted return where $\tau=\left(s_{t}, a_{t}, \ldots, s_{T}, a_{T}\right)$. The objective is to optimize the parameters $\theta$ of a policy $\pi_{{\theta}}:\mathcal{S}\times \mathcal{A} \mapsto[0,1]$ to maximize the expected return $R$, that is $\underset{\theta}{\operatorname{argmax}} \ \mathbb{E}_{\tau \sim p_{\theta}(\tau)}[R(\tau)]$
where $p_\theta$ is the trajectory distribution induced from the stochasticity of transitions, observations, and policy. In the following section we elaborate on how we integrate the \gls{rl} policy $\pi_{{\theta}}$ along with the controller to improve beyond its structure efficiently.

\subsection{Finite State Controller for Manipulation}
Following the skill formalism described in \cite{imped_meta}, we implemented a state machine for our peg-in-hole task, shown in Fig. \ref{fig:res_policies_vars}(c) and include it within our black-box controller, over which we seek improvement. This state machine controller includes five states each of which evaluates predefined success conditions $\mathcal{C}_{\mathrm{suc}}$ at run time to allow proceeding with next states or subtasks. Moreover, we use an additional state to proceed with a recovery behaviour if any of the states are not successful, i.e. evaluate to $\mathcal{C}_{\mathrm{err}}$, or the insertion sequence has finished. This state in the context of \gls{rl} serves as a reset behavior for re-initializing the environment again at test or train time. As shown in Fig. \ref{fig:res_policies_vars}(c), we have the advantage of optionally choosing which state of the state machine should improve and suggest intervention of the \gls{rl} policy only at those states, highlighted in blue. The inputs to the controller include joint position, velocity, and effort at 1KHz along with the outputs of target joint torques at the same frequency. In addition, the state machine updates set-point commands of the controller at 40Hz similar to the frequency at which residual commands are inferred from $\pi_{{\theta}}$. For each episode, this result in \emph{environment steps} of approx. 6000 while the \emph{learning steps} at which we use the RL policy (episode length) was approx. 150 which is also because we use the policy during only two out of five states of the state machine. As illustrated in Fig. \ref{fig:res_policies_vars}(c) it includes five primary states starting with "\emph{Move to pre-insert pose}" that brings the peg in a tilted orientation above a pre-defined hole position. Next, in "\emph{Find contact}", the peg is moved toward a pre-defined direction until it touches the surface after which the controller proceeds with the "\emph{search hole}" state that moves the peg on the surface while maintaining a constant vertical force on it to find the hole. Afterwards, the peg's tip is assumed to be between the hole's edges where in "\emph{Hybrid force alignment}" the robot aligns the peg with a pre-defined orientation  vertical to the surface while applying a constant force on the surface. Finally, in the "\emph{Insertion}" state, the controller, applies a vertical force to insert the peg while applying sinusoidal oscillations over the applied wrench at the end-effector frame.

\section{Hybrid Residual Feedback \& Actions}\label{sec:method}
We first define the configuration space of our robot as $C \in \mathbb{R}^7$ in addition to its joint's angular velocity $\mathcal{V}\in \mathbb{R}^7$, joint's torque space $\mathcal{T} \in \mathbb{R}^7$, end-effector pose $\mathcal{P} \in \mathbb{R}^7$ (Cartesian position and quaternion), and wrench space $\mathcal{W} \in \mathbb{R}^6$. For each modality we use superscripts "o", "u", and "$\pi$" if they represent the input to the controller, its output, and the output from sampling the policy distribution $\pi_\theta$. Accordingly, we assume $f$ is a a conventional impedance feedback controller with a mapping $f:\mathcal{C^\text{o}}\times \mathcal{V^\text{o}}\times \mathcal{T^\text{o}} \mapsto \mathcal{T^\text{u}}$.
Furthermore, we distinguish between the observations of the controller $f$ and the RL policy by $o_1 \in \mathcal{C^\text{o}}\times \mathcal{V^\text{o}} \times \mathcal{T^\text{o}}$ and $o_2 \in \mathcal{O}$ respectively. Nevertheless, the methods discussed in this section are yet agnostic to the choices of $f$, observation, and action spaces. In the following we define each sketched formulation in Fig. \ref{fig:res_policies_vars}, and Fig. \ref{fig:res_policies_expectation}(b).

\subsection{Residual Policy Learning (RPL)}
The original formulation of RPL proposes to use a residual policy whose actions are added to the outputs of the controller. However, depending on the controller's architecture, the output of the controller can be adapted at different levels of the control command, e.g., the joint effort or the end-effector wrench. Both approaches are described below. 
\subsubsection{\textbf{joint effort action}}
The original \gls{rpl} formulation, $\pi_{{\theta}}: \mathcal{O} \times \mathcal{T^\pi} \mapsto [0,1]$, proposed by \cite{res_rl} suggests superposing outputs of a \gls{rl} policy with those of the up-stream controller $f$, that is 
\begin{equation}
a_{{t}}^{rl} \sim \pi_{{\theta}}\left(\cdot \mid o_{_2{t}}\right),\quad
a_{t} = f(o_1, t) + a_{{t}}^{rl}.
\label{eq:pi_map}
\end{equation}
Here we define $\mathcal{T^\pi}$ as the action space of the policy $\pi_{{\theta}}$ which is of the same modality as the controller's outputs, in this case joint efforts. We refer to this formulation in our comparison as "joint effort action" and illustrate a schematic of it in Fig. \ref{fig:res_policies_vars}(a).

\subsubsection{\textbf{end-effector wrench}}
As mentioned earlier, although we can assume no knowledge about the input/output structure of the controller, in cases where such knowledge is available, we can certainly exploit them to our problem's advantage. In our case for \gls{rpl} we i) first map the upstream controller's output to the wrench space, ii) then superpose them with their equivalent output from the $ \pi_{{\theta}}$, and finally iii) map the result back to the original joint space. This procedure can also be done by only mapping residual wrench values to joint space, yet with less possibility for post-processing of the controller's output in wrench space. In our case, we do this conversion by $ \mathcal{F} =  \mathbf{J}^{\dagger^{T}} \tau$ and $\tau = \mathbf{J}^{T} \mathcal{F}$ where $\tau \in T$, $\mathcal{F} \in W$, $ \mathbf{J}$, and $\mathbf{J}^{\dagger}$ are joint torque commands, wrench at end-effector frame, Jacobian, and damped-pseudo inverse of the Jacobian respectively. We note this formulation as "end-effector wrench", $\pi_{{\theta}}: \mathcal{O} \times \mathcal{W^\pi} \mapsto [0,1]$, with a superposed action computed as
\begin{equation}
w_{t} \sim \pi_{{\theta}}\left(\cdot \mid o_{_2{t}}\right), \quad
a_{t} = f(o_1, t)  + \mathbf{J}^{T} w_{t}.
\label{eq:pi_map}
\end{equation}
This formulation allows using a control space that is more relevant for the task, i.e., the end-effector space, and therefore, 
intuitively, it should be easier to solve for the residual policy depending on the task. This policy design is sketched in Fig. \ref{fig:res_policies_vars}(b).

\subsection{Residual Feedback Learning (RFL)}
There is a wide range of tasks where the above formulation does not perform well as the residual policy causes a feedback distribution shift that the controller sees as external perturbation which it tries to resist. Hence, the residual actions from the RL policy results in a competition between $f$ and $\pi_{{\theta}}$. That is, as shown for example in Fig.  \ref{fig:res_policies_expectation} (b), if the $\pi_{{\theta}}$ optimal action moves the end-effector to the left, the prior-controller fights this perturbation and generates forces to retrieve the previous position. Furthermore, depending on the controller's structure, the response to such perceived external perturbation or persistent error can vary extensively. In some cases it may even lead to lower safety, especially where the residual-policy architecture lacks bounded output guarantees. To address such limitations, we propose learning a \emph{residual feedback} as an alternative to residual-action policies. Here, instead of superposing residual actions to the output of the controller, we superpose residual feedback to the feedback it receives from the environment.
\subsubsection{\textbf{joint position feedback}}
We start with a vanilla formulation where residual feedbacks are used in the original feedback space of the robot, in our case the joint positions. Our residual feedback policy can be defined as $\pi_{{\theta}}: \mathcal{O} \times \mathcal{C^\pi} \mapsto [0,1]$. The superposed action is then computed as follows
\begin{equation}
o_{_1{t}}^{rl} \sim  \pi_{{\theta}}\left(\cdot \mid o_{_2{t}}\right), \quad
a_{t} = f(o_{1} +o_{_1{t}}^{rl}, t).
\label{eq:pi_map}
\end{equation}
 For residual feedback however, we only superpose residual joint position feedback $q\in C$ \footnote{where we implicitly assume the sum $o_{_1{t}}+o_{_1{t}}^{rl}$ being a superposition of $o_{_1{t}}^{rl}$ and corresponding joints' dimensions of $o_{_1{t}}$.}, while optionally other feedback modalities can be included, e.g.. joint's angular velocities and torques, depending on the application. This policy design is sketched in Fig. \ref{fig:res_policies_vars}(d), where we only show an example of a superposition of residual joint position feedback.
\subsubsection{\textbf{end-effector pose feedback}}
We can again use a task-space centric feedback signal by converting the feedback from the environment to the task space and, after superposition with the output of the \gls{rl}-agent, map them back to their original space. For instance, we do so in our case we map joint positions to end-effector position in the base frame using forward kinematics ($\mathrm{FK}$). Afterwards we map the superposition results back using inverse-kinematics ($\mathrm{IK}$). We denote this residual feedback approach "end-effector pose feedback". In addition, this conversion can be done rapidly as the agent's outputs only make a small modification of the feedback and the optimizer that we use for inverse kinematics can use the true feedback as an initial optimization point. This procedure results in a definition of our policy $\pi_{{\theta}}: \mathcal{O} \times \mathcal{P^\pi} \mapsto [0,1]$, where $\mathcal{P^\pi}$ is the action space (i.e. residual end-effector pose), and the superposed action is computed as
\begin{equation}
\begin{aligned}
&ee = \mathrm{FK}(q_t), \quad a_{{t}}^{rl} \sim \pi_{{\theta}}(\cdot|o_{_2{t}}), \\
&q_{\text{residual}} = \mathrm{IK}(ee + a_{{t}}^{rl}), a_t = f (q_{\text{residual}}, v_t, t_t).
\end{aligned}
\label{eq:pi_map}
\end{equation}
where $ee \in \mathcal{P}^o$ and we assume the normal input to $f$ is $o_{_1{t}}=(q_t,v_t,t_t)$ (before superposing residual feedback). This policy design is sketched in Fig. \ref{fig:res_policies_vars}(e). The formulation again allows adapting the feedback in a space that is more relevant for the task.

\subsection{Hybrid Residual Reinforcement Learning (HRRL)}
Each formulation of residual reinforcement learning and feedback learning has advantages depending on the stage of the task execution (cf. our experiments in section \ref{sec:contact_based_exp}). This allows, for example, sudden actions or high frequency vibrations that are needed for releasing a peg that is stuck due to overcome orientation uncertainty with residual actions as well as more flexible spatial movement of the peg in search for the hole with residual feedback. For this reason we propose a combination of residual action and residual feedback to form a new residual hybrid model, called "joint space hybrid". We illustrate a schematic of our hybrid model in Fig. \ref{fig:res_policies_expectation}(a). In our case, we extend the action space to 14 dimensions where the first 7 dimensions contribute within the original residual policy formulation and the remaining dimensions modify the feedback. In particular, our hybrid residual policy is defined as $\pi_{{\theta}}: \mathcal{O} \times \mathcal{T^\pi} \times \mathcal{C^\pi} \mapsto [0,1]$, with a superposition,
\begin{equation}
a_{{t}}^{rl}, o_{_1{t}}^{rl} \sim  \pi_{{\theta}}\left(\cdot \mid o_{_2{t}}\right),\quad
a_{t} =  f(o_{_1{t}} +o_{_1{t}}^{rl},t) + a_{{t}}^{rl}.
\label{eq:pi_map}
\end{equation}
Note that $o_{_1{t}}^{rl} \in \mathcal{C^\pi}$ only represents residual feedback of the joint modality. For simplicity we implicitly assume the sum $o_{_1{t}}+o_{_1{t}}^{rl}$ is a superposition of $o_{_1{t}}^{rl}$ and corresponding the joints' dimensions of $o_{_1{t}}$. 

\section{Experiments}
%

In the following we evaluate the above formulations experimentally to observe their advantages and disadvantages within different problem settings. This includes comparing them in terms of final-performance, sample efficiency, as well as well as an analysis of the benefits of each method over the others. While denoting our experimental settings, we start with simulations first and illustrate our result in the real environment in the end. 

\subsection{Workspace}

\begin{figure}
\centering
        \subfloat[Real scene]{%
                    {\includegraphics[height=0.145\textheight]{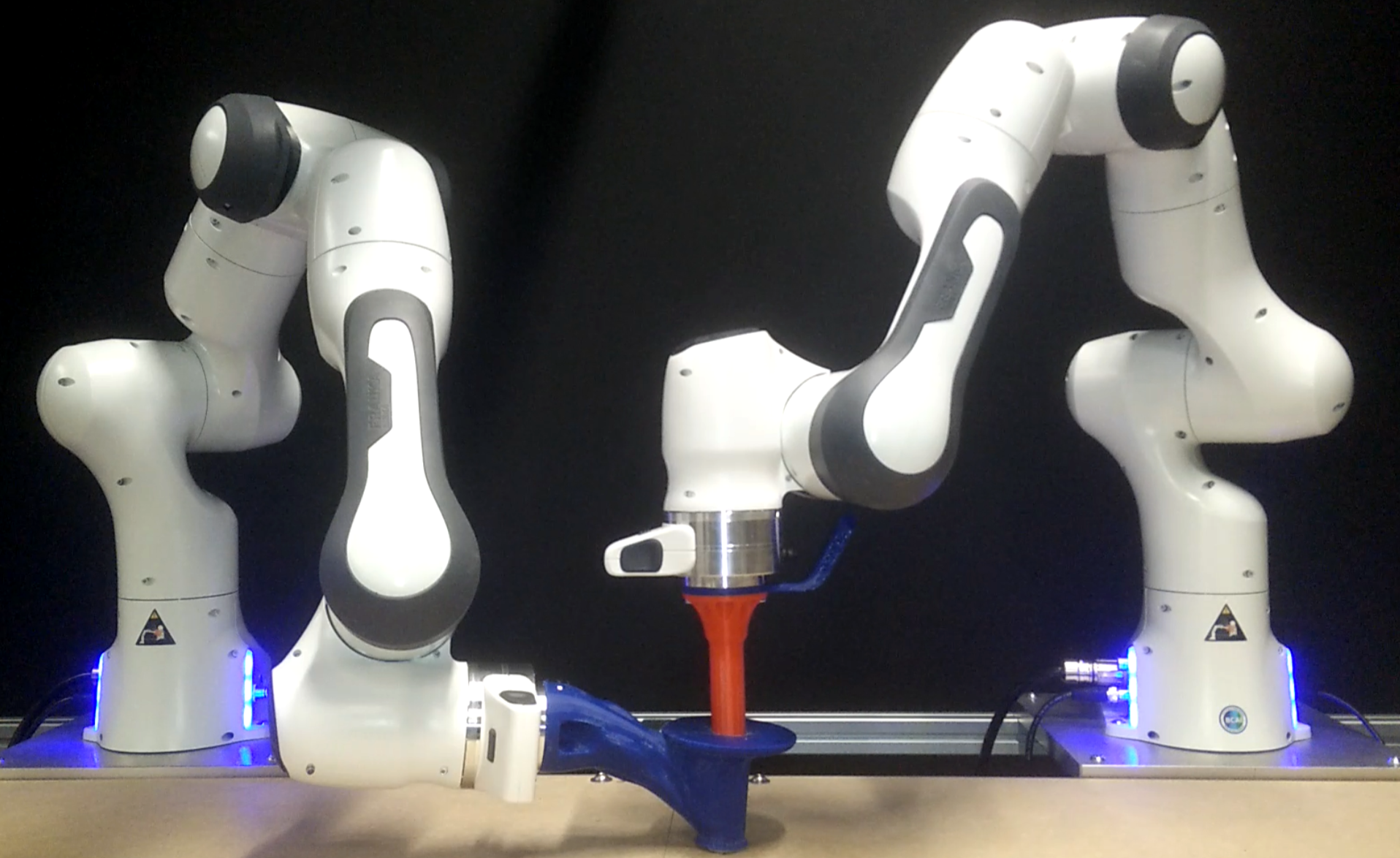}}
                    }
        \subfloat[Simulation scene]{%
                    {\includegraphics[height=0.145\textheight]{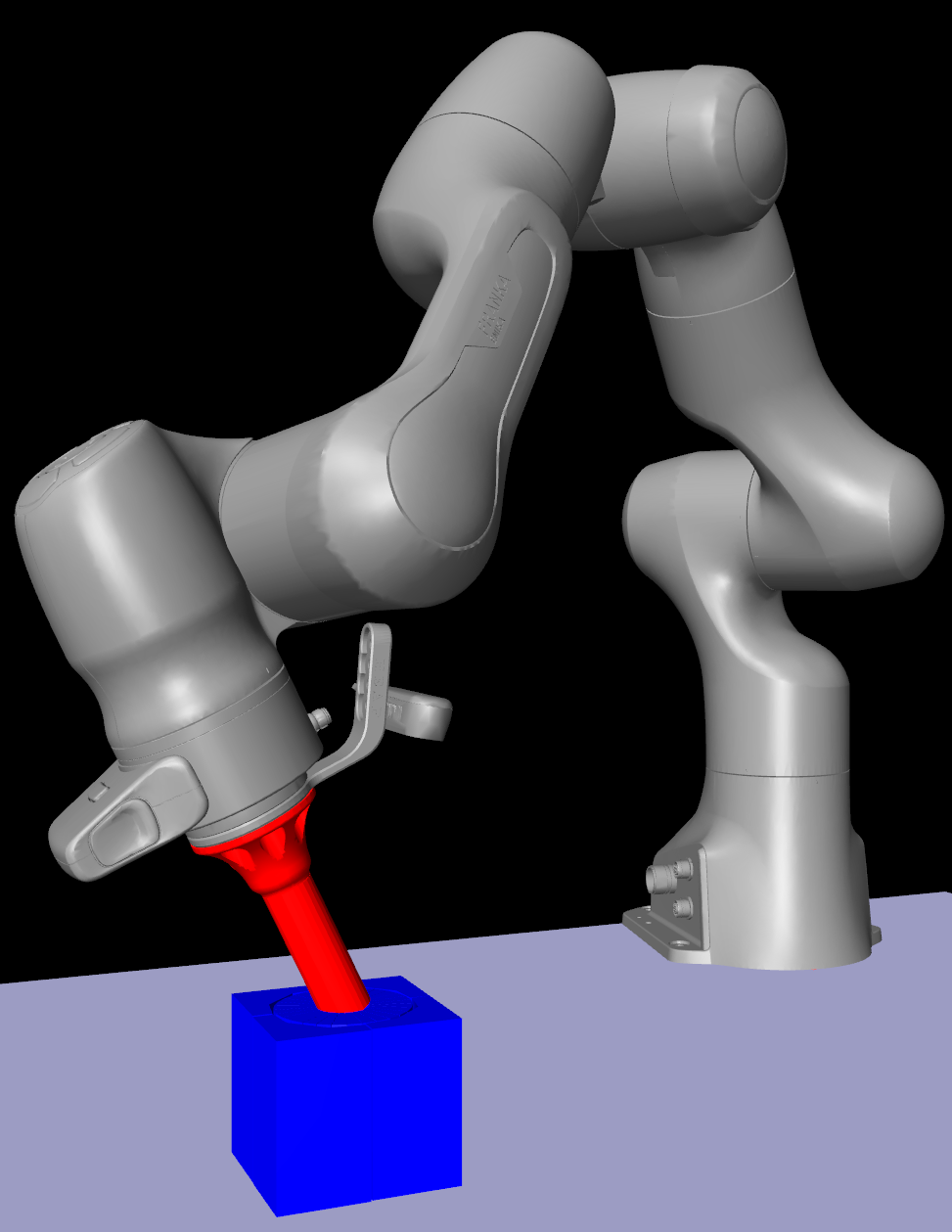}}
                    }
\caption[Real hardware workspace]{Franka Emika robots at train time for peg-in-the-hole task. While one robot is responsible for insertion of the peg, the other robot serves as a mechanism to introduce uncertainties of the environment.}
\label{fig:twin_pand}
\end{figure}

We base our experiments on a peg-in-the-hole task where a shaft of 70mm length and 25mm diameter is used as the peg. We use the Franka Emika robot arm with seven \gls{dof} to fully insert this peg in a hole of 25.8mm diameter. Two robot arms were used --- one for peg insertion while the second arm rearranged the hole between each episode for simulating the pose and orientation uncertainty. To leverage quick experiments while focusing on the difficulty of contact rich insertion rather than grasping, we fixated the shaft and hole designs to the robots' arm as shown in Fig. \ref{fig:twin_pand}. 
Compared to some of the earlier peg insertion works, e.g. \cite{ ding2019transferable,lee2019making,schoettler2020metareinforcement,hoppe_20} our task is more difficult due to the larger size of the peg in terms of diameter, length, and raw surface roughness owing to the 3D-printer's outlet. The end-effector commands and state readings are computed with respect to the \gls{tcp} at the tip of the peg. For our \gls{rl} learner we use the PyTorch \gls{ppo} implementation from \cite{pytorchrl_ppo}.

We introduce uncertainty to our task in terms of the position and orientation of the hole in meters and radians respectively, i.e., before each episode, the hole position and orientation are sampled from a Gaussian distribution and the second arm rearranges the hole accordingly. The variance of this Gaussian distribution directly relates to the task difficulty and is specified by the curriculum. The position of the hole is not known to the second robot (the learning agent). In addition, to allow evaluation and debugging in simulations for orientation uncertainty separate from that of the position, the orientations are calculated with respect to a central point on the upper side of the hole to allow the peg's tip to always fall down between the hole edges. In all training experiments we choose the sparse reward of $r(s, a)$ which is $1 \text { if } \left\|P_{tcp}  - P^{*}_{tcp}\right\|_{2} < \epsilon$, and 0 $\text {otherwise }$, where $P_{tcp}$ and $P^{*}_{tcp}$ are the current and goal \gls{tcp} positions. The value of $P^{*}_{tcp}$ is computed based on the lower side of the hole's position and $\epsilon$ was set to 5mm. We initialize  the last layer of the policies with zero weights such that we start with a plain execution of the prior-controller. We use the first 50 episodes to only learn the critic without updating the policy. The starting points of all curves are regardless of the chosen algorithm.

We use MuJoCo \cite{Mujoco_12} for simulation of our peg-in-the-hole assembly task. Only one robot is simulated as the hole can be rearranged programmatically. Since the physics engines require all simulated objects to be convex, we automated the design of CAD models to fulfill this requirement  using a python script in Blender \cite{blender_sim} that decomposes separate mesh parts for MuJoCo. A simulation scene can be found in Fig. \ref{fig:twin_pand}.
\subsection{Searching for Hole using Visual Inputs}\label{sec:contact_exp}

Although our focus regarding industrial assembly is mainly on using contact feedback, we find using visual inputs while moving towards the hole in the air better to contrast the advantage of using residual feedback compared to residual actions. We do so during the "Move to pre-insert pose" state of the state machine, shown in Fig. \ref{fig:res_policies_vars} (c) and consider 1.6 centimeters position uncertainty for the hole without any orientation uncertainty. For our observations $\mathcal O$ we choose 84x84x3 RGB images from a hand-mounted camera, as well as the robot's end-effector's Cartesian position and orientation in Euler angles along with its wrench. In addition, we choose a convolutional policy architecture similar to \cite{wu2017scalable}. These convolutional layers that receive RGB inputs comprise of 32 filters of 8x8 size with stride 4 followed by 64 filters of size 4x4 with stride 2, and 32 filters of size 3x3 with single stride, all of which use ReLU activation. We concatenate the output of the convolutional layers with a latent representation of the robot's state similar to \cite{lukas} before the subsequent actor's and critic's fully connected layers.

\begin{figure}
\centering
\scalebox{0.63}{\includegraphics{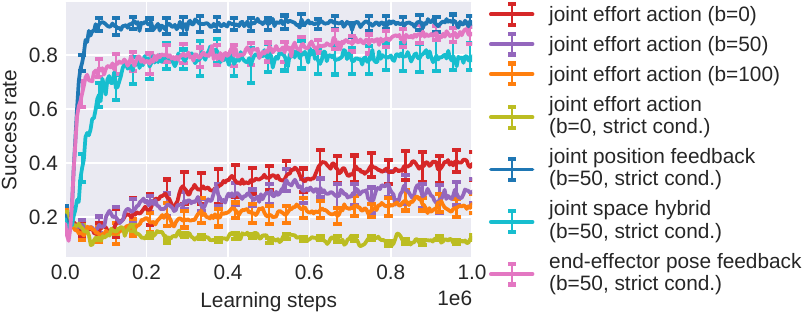}}
\caption[Residual feedback learning formalism]{Performance of using residual feedback learning compared to the original residual policy learning. The environment difficulty includes only 1.6 (cm) position uncertainty. Error bars correspond to half of the standard deviation over 30 seeds.}
\label{fig:rl_eefb_baseline}
\end{figure}

Here, with the original formulation of \gls{rpl} (e.g., "joint effort action" in our case), ideal residual actions try to move the peg above the hole while the upstream controller observes this interaction as external perturbation and resists it. This effect becomes more damaging when the frequency of the controller is higher than the frequency of the \gls{rl}-agent interventions. That is, if we update each residual action after every 10 controller steps, those 10 steps represent opportunities to counteract the external perturbation and hence compete with the \gls{rl}-policy. One way to investigate such an issue is using different \gls{rl}-agent frequencies. However, this would also change the episode length and the learning problem. For this reason, we only add a number of controller "buffer steps" during which the controller works without any update from the RL policy, giving the aforementioned opportunity to recover from any external perturbations, such as those from the RL policy. We add these buffer steps at the end of the "Move to pre-insert pose" each of which correspond to 1 millisecond of using the controller without any RL policy inference. Furthermore, regardless of the sensitivity of the low-level control layers to their feedback (e.g., impedance controller in our case), higher level layers such as state machines or a behavior trees can also respond critically to the feedback. We show an example of such case that we refer to it as "strict condition" in our comparison in Fig.  \ref{fig:rl_eefb_baseline}, where the state machine raises an error if it does not observe achieving its goal that can be the result of any external perturbation such as those coming from the residual action policy. As illustrated in this figure, higher buffer steps, e.g., b=100, gives the up-stream controller more opportunity to resist the residual policy, resulting in lower success rates. In contrast, using residual feedback does not result in any competition between the \gls{rl} policy and the low level controller, while also keeping the higher level state machine satisfied in achieving its goal. We observe these results from the {\bf joint position feedback} baseline in Fig. \ref{fig:rl_eefb_baseline}.


\subsection{Using contact and proprioceptive Features as Inputs} \label{sec:contact_based_exp}
\begin{figure}
    \centering
    \includegraphics[width=0.48\textwidth]{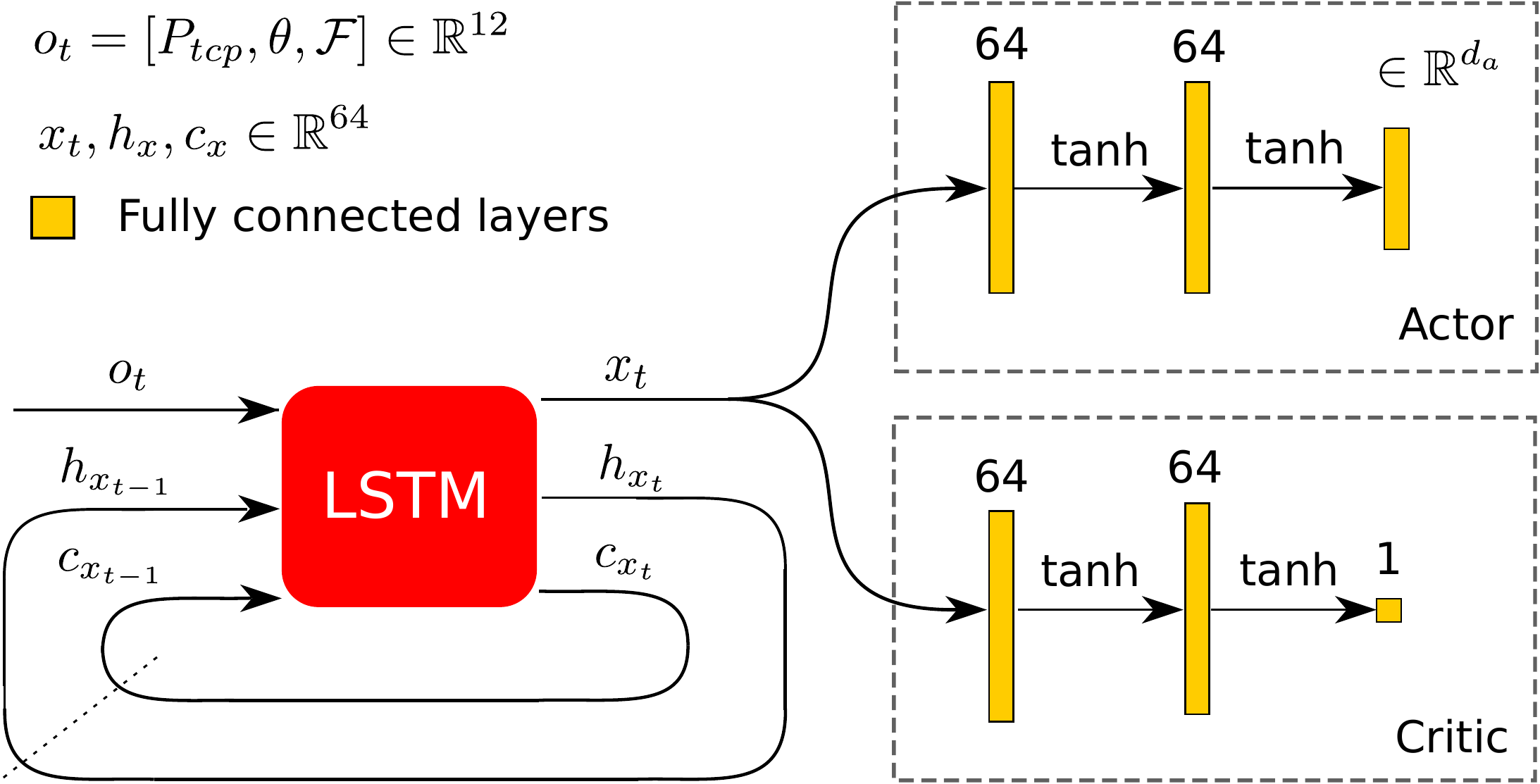}
    \caption[Network Architectures]{The network architecture used in Section \ref{sec:contact_based_exp}. We initialized our policy architecture with (semi) orthogonal  matrices of gain $\sqrt{2}$, except the last layer of the actor which similar to \cite{silver2018residual} is zero initialized.}%
    \label{fig:architecture1}%
\end{figure}

\begin{figure*}
\centering

\subfloat{%
            {\includegraphics[width=0.99\textwidth]{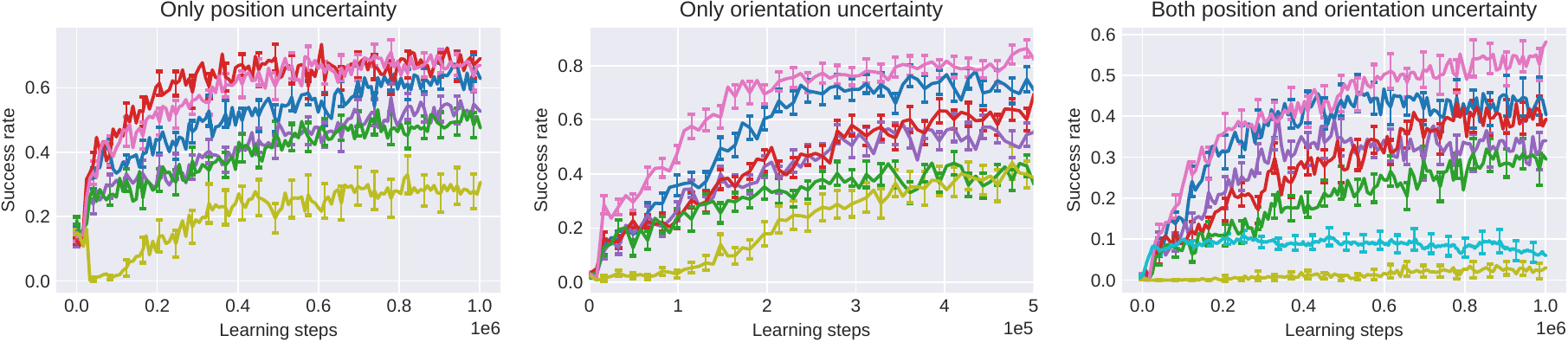}}
            }
\qquad   
\subfloat{%
            {\includegraphics[width=0.99\textwidth]{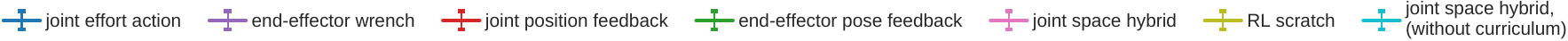}}
            }
\caption[baselines comparison]{Evaluation of each residual policy variant in simulation averaged over 40 seeds using 8 parallel environments four of which are for training (with variable difficulty)  and the rest for evaluation (with constant difficulty). The constant difficulty of these evaluations are noted in Table \ref{tab:curr_params}. The starting point of each curve correspond the zero-shot evaluation of controller for 50 episodes. Error bars correspond to a quarter of the standard deviation.}
\label{fig:baselines_comparison}
\end{figure*}

\begin{figure}
    \centering
    \subfloat{%
                    {\includegraphics[width=0.23\textwidth]{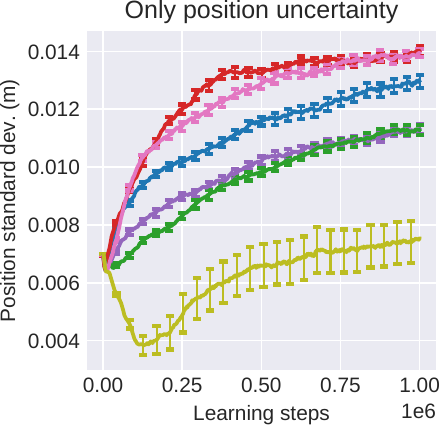}}
                }
    \subfloat{%
                {\includegraphics[width=0.23\textwidth]{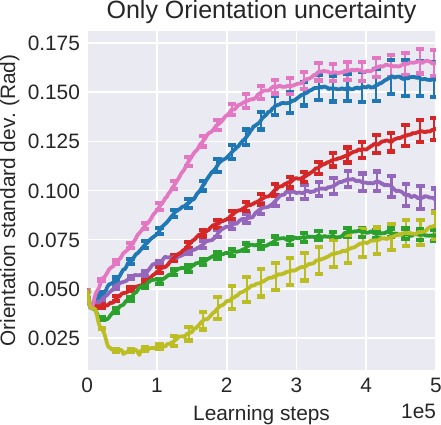}}
                }%
    \vspace{1pt}
    \qquad   
    \subfloat{%
                {\includegraphics[width=0.47\textwidth]{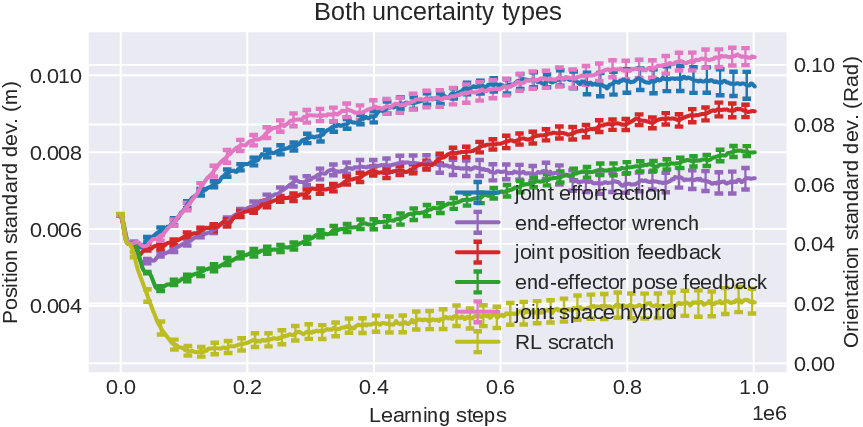}}
                }                    
    \caption[Simulation Scene]{History of the domain randomization parameters, i.e., the curriculum difficulty at train-time. Error bars correspond to a quarter of the standard deviation.}%
    \label{fig:curr_hist}%
\end{figure}

In this experiment, the peg-in-the-hole task needs to be learned without vision feedback, only relying on the proprioceptive feedback  and the contact force readings at the joints. We choose the relative position of the end-effector from the position where the first contact takes place denoted as $P_{tcp}$, along with its orientation in euler angels, $\theta$, and the measured contact wrench $\mathcal{F}$ at the end-effector frame. For the policy architecture, we use \gls{lstm} similar to \cite{Inoue_lstm_peg_17the}, 
which is shared between the actor and the critic as shown in Fig. \ref{fig:architecture1}. The actor computes the mean of a $d_a$ dimensional Gaussian distribution from which actions are sampled, (see Section \ref{sec:method} for a description of the action definitions).
Here, achieving a successful insertion includes solving two main sub-tasks that are i) search for the hole by moving the peg's tip on the surface, and ii) insertion, during which orientation uncertainty is of challenge. We aim for improving the prior controller in obtaining this goal by allowing the \gls{rl} policy to intervene during relevant sub-tasks as shown in Fig. \ref{fig:res_policies_vars}(c). In addition, for training our policy we leverage the adaptive curriculum formulation from \cite{lukas} with the difference that our environments adapt the difficulty independent of each other. This curriculum increases or decreases the degree of domain randomization (variance in the hole position and orientation) depending on the current success rate of the policy which helps us significantly on sparse-reward domains like our current setting. That is, when the current success rate is below 0.6, we decrease the difficulty, and increase if the success rate is above 0.7. We list the parameters of such adaptive domain randomization as well as the uncertainties involved in our evaluation environments in Table \ref{tab:curr_params}. Moreover, Fig. \ref{fig:curr_hist} shows the history of this adaptation in our experiments and Fig. \ref{fig:baselines_comparison} illustrates the results of our evaluations concurrent with training. These evaluations were done for position and orientation uncertainty separately as well as together. Additionally, we include a comparison with learning from scratch where the controller is only used for finding contact and alignment of the peg while the rest is purely controlled by an \gls{rl} agent using joint torques as the action space. In addition, we also experiment the "joint space hybrid" formulation that had shown superior performance but without the adaptive curriculum. This baseline that we denote "joint space hybrid (without curriculum)" in Fig. \ref{fig:baselines_comparison} trains the environments with the same difficulty at which the other baselines are evaluated, and due to this high difficulty, it rarely observe any reward to learn. 

\begin{table}[b]
\centering
\begin{tabular}{c|l|c|c|c}
\multicolumn{2}{c|}{\multirow{2}{*}{\begin{tabular}[c]{@{}c@{}}Adaptive \\domain randomization\\~parameters\end{tabular}}}                                         & \multicolumn{3}{c}{Experiment}                                                                                                                                                 \\ 
\cline{3-5}
\multicolumn{2}{c|}{}                                                                                                                                              & \begin{tabular}[c]{@{}c@{}}Only\\Position\end{tabular} & \begin{tabular}[c]{@{}c@{}}Only\\Orienration\end{tabular} & \begin{tabular}[c]{@{}c@{}}Both\\Unc. types\end{tabular}  \\ 
\hline
\multirow{3}{*}{\begin{tabular}[c]{@{}c@{}}\textcolor[rgb]{0.2,0.2,0.2}{Position std.}\\\textcolor[rgb]{0.2,0.2,0.2}{(Meters)}\end{tabular}} & Initial              & 0.007                                                  & 0                                                         & 0.007                                                     \\ 
\cline{2-5}
                                                                                                                                            & Evaluation           & 0.016                                                  & 0                                                         & 0.015                                                     \\ 
\cline{2-5}
                                                                                                                                            & Increment            & 0.001                                                  & 0                                                         & 0.001                                                     \\ 
\hline
\multirow{3}{*}{\begin{tabular}[c]{@{}c@{}}Orientation std.\\(Radians)\end{tabular}}                                                         & Initial              & 0                                                      & 0.05                                                      & 0.05                                                      \\ 
\cline{2-5}
                                                                                                                                            & Evaluation           & 0                                                      & 0.15                                                      & 0.1                                                       \\ 
\cline{2-5}
                                                                                                                                            & Increment            & 0                                                      & 0.01                                                      & 0.01                                                      \\ 
\hline
\multicolumn{2}{c|}{success rate bounds}                                                                                                                           & {[}0.6, 0.7]                                           & {[}0.6, 0.7]                                              & {[}0.6, 0.7]                                              \\
\multicolumn{1}{l}{}                                                                                                                        & \multicolumn{1}{l}{} & \multicolumn{1}{l}{}                                   & \multicolumn{1}{l}{}                                      & \multicolumn{1}{l}{}                                     
\end{tabular}
\caption{Adaptive curriculum parameters used in each experiment that we describe in section \ref{sec:contact_exp}. The evaluation values correspond the constant difficulty at which the policies were evaluated in Fig. \ref{fig:baselines_comparison} concurrent with training.} 
\label{tab:curr_params}
\end{table}


As the results shown in Fig. \ref{fig:baselines_comparison} suggest,  \textbf{joint position feedback} appears superior for handling position uncertainty, \textbf{joint effort action} for orientation uncertainty, while mitigating both uncertainty types could be achieved best with \textbf{joint space hybrid}. We see the same observations from Fig. \ref{fig:curr_hist} where \textbf{joint space hybrid} can adapt to the widest range of uncertainties, i.e. always maintain a high success rate on the largest standard deviations of domain parameters. This makes sense as overcoming position uncertainty requires more spatial movement of the peg which is difficult with "joint effort action" (i.e., \gls{rpl}) due to the resistance of the controller. In contrast, for orientation uncertainty, behaviors such as a rather high frequency vibration of the peg in the hole help pushing the peg into the hole. Yet, this may not be possible with \gls{rfl} learning as the controller is often designed to produce smooth output actions. In this case, the \gls{rpl} can, for example, conduct non smooth or sudden actions through the residual outputs that are directly applied to the environment. From our results, it can be seen that our hybrid residual RL framework combines the strength of both approaches and shows superior performance in the single pose or orientation uncertainty tasks as well as the complete task.

\subsection{Hardware experiments}
\begin{figure}[h]
    \centering
    \includegraphics[width=0.47\textwidth]{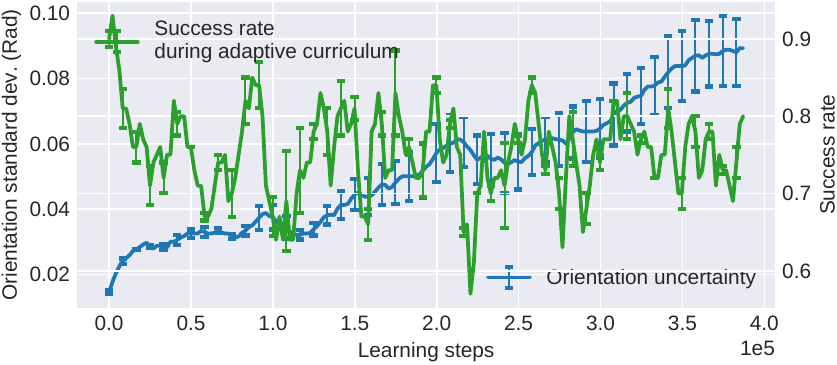}
    \caption{Adapted orientation uncertainty based on curriculum as well as success rate of the training environment on hardware over two different seeds. Error bars include half of standard deviation around the mean.}%
    \label{fig:hw_R_curr}%
\end{figure}

We trained our policy on hardware with a speed of approx. 3 episodes per minute with the same observation space we described in section \ref{sec:contact_based_exp} and only using our "joint space hybrid" formulation. As we also leverage adaptive domain randomization here, the evaluation of the policy concurrent with training similar to the simulations would have doubled our experiment time. For this we use our real environment only for training and evaluate it afterwards.  We illustrate the history of the adaptation in Fig. \ref{fig:hw_R_curr} as well as success rate at train time. Every time the environment's difficulty increases with respect to the curriculum the observed success rate decreases as well that is the reason for the oscillatory shape of the learning curve. Our training starts with  position and orientation uncertainty of 5 millimeters and 0.015 Radian. In addition, our adaptive curriculum only modifies the orientation uncertainty using the second robot as shown in Fig. \ref{fig:twin_pand}(a), by an increment of 0.0025 Radian depending on the success rate over 15 episodes. Using this adaptive formulation offered us the convenience of not having to predefine the difficulty of the environment without knowing if it is too high or low while  automatically increasing the degree of domain randomization on hardware. One single experiment in total took 12 hours. Finally, our evaluations demonstrated the success rate of 0.92 within 25 trials orientation and uncertainty of 0.08 Radians. A  video  showing  the  results  can  be found at 
 \href{https://youtu.be/SAZm_Krze7U}{\texttt{https://youtu.be/SAZm{\char`_}Krze7U}.}

\section{CONCLUSION AND FUTURE WORK}
Every controller comes with numerous imperfections that limit its performance. Hence, it is desirable to improve beyond their underlying structures using least engineering effort and in a sample efficient way. We demonstrated the limitations of the existing \gls{rpl} formulation for controller improvement that uses residual actions and proposed a more flexible extension that can also exploit residual feedback. As we demonstrated, both residual formulations have distinct advantages that can be exploited in different task settings. We demonstrated leveraging both residual models simultaneously within a hybrid model and gained superior performance over all cases. In contrast to the \gls{rpl} formulation where the up-stream controller resists the external perturbation caused by the \gls{rl} policy, the \gls{rfl} applies changes in the feedback signal, which can be regarded as similar to changing the set-point of the controller in many cases. 

Furthermore, we chose industrial assembly as a good example of a scenario where lack of accurate models to represent contact rich dynamics is of great challenge especially if uncertainties need to be considered. While we demonstrate sample efficient improvement of an assembly task example, i.e., peg-in-hole using raw sensory inputs, using learned latent representations of those inputs should promise even more efficiency and generalization within \gls{pomdp} settings. This applies also to different choices of \gls{rl} algorithms that are more sample efficient than PPO as our formulations are agnostic such a choice.

Finally, sim-to-real transfer of policies that only observe contact inputs is a major challenge that we kept for future work. In addition, one can also investigate better exploration strategies as well as more guarantees regarding residual policy formulations. As we took the advantage of having another robot side by side with the main robot responsible for our task, an interesting future extension is to have two robots competing with each other in an adversarial setting where one robot tries to learn successful insertions while the other robot learns to make insertions more difficult.


\addtolength{\textheight}{-12cm}   







\bibliographystyle{IEEEtran}
\bibliography{root}

\end{document}